\documentclass{llncs}

\usepackage{graphicx}

\usepackage{subfigure}
\usepackage{amsmath}
\usepackage{amssymb}
\usepackage{wrapfig}
\usepackage{upgreek}
\usepackage{multirow}

\begin{document}
\title{Variational AutoEncoder For Regression: Application to Brain Aging Analysis}
\author{Qingyu Zhao\inst{1}, Ehsan Adeli\inst{1}, Nicolas Honnorat\inst{2}, Tuo Leng\inst{1}, Kilian M. Pohl\inst{1,2}}
\institute{Stanford University, Stanford, CA, USA \and SRI International, Menlo Park, CA, USA}
\maketitle
\begin{abstract}
While unsupervised variational autoencoders (VAE) have become a powerful tool in neuroimage analysis, their application to supervised learning is under-explored. We aim to close this gap by  proposing a unified probabilistic model for learning the latent space of imaging data and performing supervised regression. Based on recent advances in learning disentangled representations, the novel generative process explicitly models the conditional distribution of latent representations with respect to the regression target variable. Performing a variational inference procedure on this model leads to joint regularization between the VAE and a neural-network regressor. In predicting the age of 245 subjects from their structural Magnetic Resonance (MR) images, our model is more accurate than state-of-the-art methods when applied to either region-of-interest (ROI) measurements or raw 3D volume images. More importantly, unlike simple feed-forward neural-networks, disentanglement of age in latent representations allows for intuitive interpretation of the structural developmental patterns of the human brain.
\end{abstract}

\section{Introduction}
Generative models in combination with neural networks, such as \textit{variational autoencoders} (VAE), are often used to learn complex distributions underlying imaging data \cite{Benou16}. VAE assumes each training sample is generated from a latent representation, which is sampled from a prior Gaussian distribution through a neural-network, i.e., a decoder. Inferring the network parameters involves a variational procedure leading to an encoder network, which aims to find the posterior distribution of each training sample in the latent space. As an unsupervised learning framework, VAE has successfully been applied to several problems in neuroimaging, such as denoising \cite{Benou16}, abnormality detection \cite{Baur18} or clustering tasks \cite{Zhao19}. However, the use of VAE is still under-explored in the context of supervised regression; i.e., regression aims to predict a scalar outcome from an image based on a given set of training pairs. For instance in neuroimage analysis, the scalar could be a binary variable indicating if a subject belongs to the control or a disease group or a continuous variable encoding the age of a subject.  

Several attempts have been made to integrate regression models into the VAE framework by directly performing regression analysis on the latent representations learned by the encoder \cite{Yoo17,Chen18}. These works, however, still segregate the regression model from the autoencoder in a way that the regression needs to be trained by a separate objective function. To close the gap between the two models, we leverage recent advances in learning \textit{disentangled latent representations} \cite{Higgins17,kim18}. In the latent space, a representation is considered disentangled if changes along one dimension of that space are explained by a specific factor of variation (e.g., age), while being relatively invariant to other factors (e.g., sex, race) \cite{Higgins17}. Herein, we adopt a similar notion to define a unified model combining regression and autoencoding. We then test the model for predicting the age of a subject solely based on its structural MR image. 

Unlike a traditional VAE relying on a single latent Gaussian to capture all the variance in brain appearance, our novel generative age-predictor explicitly formulates the conditional distribution of latent representations on age while being agnostic to the other variables. Inference of model parameters leads to a combination between a traditional VAE network that models latent representations of brain images, and a regressor network that aims to predict age. Unlike the traditional VAE, our model is able to disentangle a specific dimension from the latent space such that traversing along that dimension leads to age-specific distribution of latent representations. We show that through this mechanism the VAE and the regressor networks regularize each other during the training process to achieve more accurate age prediction.

Next, we introduce the proposed VAE-based regression model in Section 2. Section 3 describes the experiments of age prediction for 245 healthy subjects based on their structural T1-weighted MR images. We implement the model using two network architectures: a multi-layer perception for imaging measurements and a convolutional neural network for 3D volume images. Both implementations achieve more accurate predictions compared to several traditional methods. Finally, we show that the learned age-disentangled generative model provides an intuitive interpretation and visualization of the developmental pattern in brain appearance, which is an essential yet challenging task in most existing deep learning frameworks.

\section{VAE for Regression}
\begin{figure}[!t]
	\centering
    \includegraphics[width=0.28\linewidth]{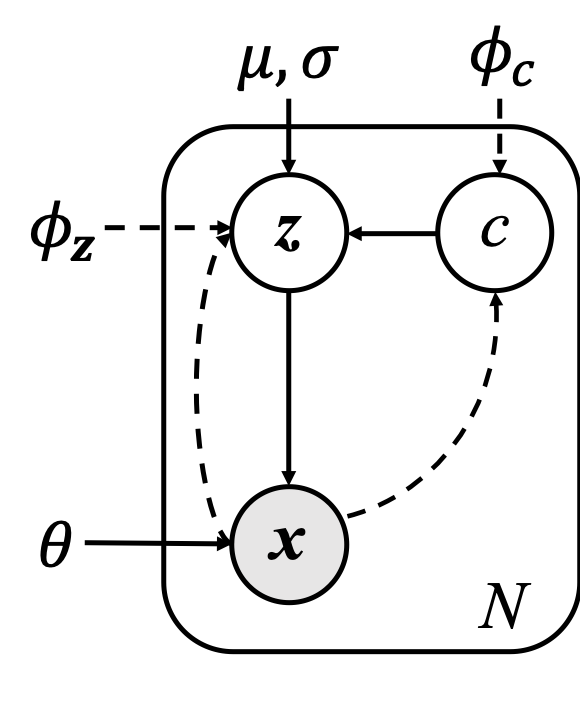}
    \includegraphics[width=0.55\linewidth]{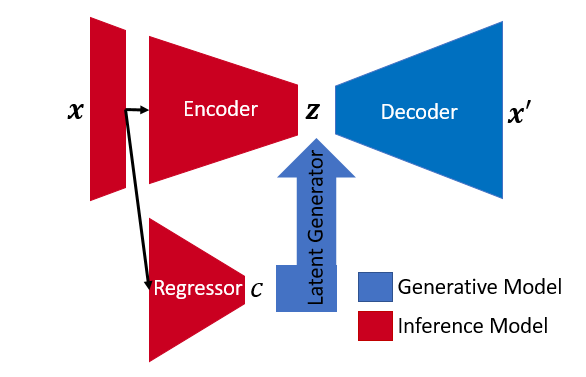}
    \caption{Probabilistic (left) and graphical (right) diagrams of the VAE-based regression model. Each image $\boldsymbol{x}$ is assumed to be generated from its representation $\boldsymbol{z}$, which is dependent on age $c$ (blue blocks). The inference model (red blocks) constructs a probabilistic encoder for determining the latent representation and a probabilistic regressor for predicting age.}
	\label{fig:formulation}
\end{figure}

Fig. \ref{fig:formulation} provides an overview of the model with blue blocks representing the generative model and red blocks the inference model. 

\textbf{The Generative Model.} Let $\textbf{X}=\{\boldsymbol{x}^{(1)},...,\boldsymbol{x}^{(n)}\}$ be a training dataset containing structural 3D MR images of $n$ subjects, and $\textbf{C}=\{\boldsymbol{c}^{(1)},...,\boldsymbol{c}^{(n)}\}$ be their age. We assume each MR image $\boldsymbol{x}$ is associated with a latent representation $\boldsymbol{z} \in \mathbb{R}^M$, which is dependent on $c$. Then the likelihood distribution underlying each training image $\boldsymbol{x}$ is $p(\boldsymbol{x})=\int_{\boldsymbol{z},c}p(\boldsymbol{x},\boldsymbol{z},c)$, and the generative process of $\boldsymbol{x}$ reads $p(\boldsymbol{x},\boldsymbol{z},\boldsymbol{c}) = p(\boldsymbol{x}|\boldsymbol{z})p(\boldsymbol{z}|\boldsymbol{c})p(c)$, where $p(c)$ is a prior on age. In a standard VAE setting \cite{Kingma13}, the `decoder' $p(\boldsymbol{x}|\boldsymbol{z})$ is parameterized by a neural network $f$ with the generative parameters $\theta$, i.e., $p(\boldsymbol{x}|\boldsymbol{z}) \sim \mathcal{N}(\boldsymbol{x};f(\boldsymbol{z};{\theta}),\textbf{I})$ \footnote{when $\boldsymbol{x}$ is binary, a Bernoulli distribution can define $p(\boldsymbol{x}|\boldsymbol{z}) \sim \mbox{Ber}(\boldsymbol{x};f(\boldsymbol{z};\theta))$}. 
Different from the traditional VAE is the modeling of latent representations. Instead of using a single Gaussian prior to generate $\boldsymbol{z}$, we explicitly condition $\boldsymbol{z}$ on age $c$, such that the conditional distribution $p(\boldsymbol{z}|c)$ captures an age-specific prior on latent representations. We call $p(\boldsymbol{z}|c)$ a \textit{latent generator}, from which one can sample latent representations for a given age. We further assume the non-linearity of this generative model can be fully captured by the decoder network $p(\boldsymbol{x}|\boldsymbol{z})$, such that a linear model would suffice to parameterize the generator: $p(\boldsymbol{z}|c) \sim \mathcal{N}(\boldsymbol{z};\boldsymbol{u}^{\textbf{T}}c,\sigma^2\textbf{I})$, $\boldsymbol{u}^{\textbf{T}}\boldsymbol{u}=1$. With this construction we can see that $\boldsymbol{u}$ is essentially the disentangled dimension \cite{Higgins17} associated with age; traversing along $\boldsymbol{u}$ yields age-specific latent representations. Note this model does not reduce the latent space to 1D but rather links one dimension of the space to age.

\noindent\textbf{Inference Procedure.} The parameters of the above generative model can be estimated via maximum likelihood estimation (MLE)  i.e.,  by maximizing the sum of log likelihood $\sum_{i=1}^N \log p(\boldsymbol{x}^{(i)})$. For such an optimization, we adopt a standard procedure of variational inference and introduce an auxiliary function $q(\boldsymbol{z}^{(i)},c^{(i)}|\boldsymbol{x}^{(i)})$ to approximate the true posterior $p(\boldsymbol{z}^{(i)},c^{(i)}|\boldsymbol{x}^{(i)})$. In the following we omit index $i$ for convenience. In so doing,  $\log p(\boldsymbol{x})$ can be rewritten as the sum of the KL-divergence $D_{KL}$ between $q(\boldsymbol{z},c|\boldsymbol{x})$ and $p(\boldsymbol{z},c|\boldsymbol{x})$ and the `variational lower-bound' $\mathcal{L}(\boldsymbol{x})$:   
\begin{equation}
    \log p(\boldsymbol{x}) = D_{KL}\left(q(\boldsymbol{z},c|\boldsymbol{x}\right) ~||~ p(\boldsymbol{z},c|\boldsymbol{x})) + \mathcal{L}(\boldsymbol{x}).    \label{eq:mle}
\end{equation}
Based on the mean-field theory, we further assume $q(\boldsymbol{z},c|\boldsymbol{x})=q(\boldsymbol{z}|\boldsymbol{x})q(c|\boldsymbol{x})$. Then the lower-bound can be derived as 
\newcommand\numberthis{\addtocounter{equation}{1}\tag{\theequation}}
\begin{align*}
\mathcal{L}(\boldsymbol{x}):=&-D_{KL}\left(q(c|\boldsymbol{x})~||~p(c)\right)\\&+\mathbb{E}_{q(\boldsymbol{z}|\boldsymbol{x})}\left[\log p(\boldsymbol{x}|\boldsymbol{z}) \right]-\mathbb{E}_{q(c|\boldsymbol{x})}\left[D_{KL}\left(q(\boldsymbol{z}|\boldsymbol{x}\right)~||~p(\boldsymbol{z}|c))\right] \numberthis
\label{eq:vlb}
\end{align*}

In the above equation, we formulate $q(c|\boldsymbol{x})$ as a univariate Gaussian $q(c|\boldsymbol{x})\sim\mathcal{N}(c;f(\boldsymbol{x};{\phi}_c),g(\boldsymbol{x};{\phi}_c)^2)$, where $\phi_c$ are the parameters of the inference networks. We can see that $q(c|\boldsymbol{x})$ is essentially a regular feed-forward regression network with an additional output being the uncertainty (i.e., standard deviation) of the prediction. In this work we call $q(c|\boldsymbol{x})$ a \textit{probabilistic regressor}. In an unsupervised setting, the KL-divergence of Eq. \eqref{eq:vlb} regularizes the prediction of $c$ with a prior. However, in our supervised setting this term can be simply replaced by $\log q(c|\boldsymbol{x})$ as  the ground-truth of $c$ is known for each training sample \cite{Kingma14,Nalisnick17}\footnote{In a semi-supervised setting where no informative prior in present, $\mathbb{H}(q(c|\boldsymbol{x}))$, i.e., the entropy of $q(c|\boldsymbol{x})$, is commonly used to replace the last term of Eq. 4 for samples with unknown $c$ \cite{Kingma14,Nalisnick17}.}.

Similar to a traditional VAE, the remaining part of the inference involves the construction of a \textit{probabilistic encoder} $q(\boldsymbol{z}|\boldsymbol{x})$, which maps the input image $\boldsymbol{x}$ to a posterior multivariate Gaussian distribution in the latent space $q(\boldsymbol{z}|\boldsymbol{x})\sim \mathcal{N}(\boldsymbol{z};f(\boldsymbol{x};{\phi}_z),g(\boldsymbol{x};{\phi}_z)^2\textbf{I})$. Then the second term of Eq. \eqref{eq:vlb} encourages the decoded reconstruction from the latent representation to resemble the input \cite{Kingma13}. The third term of of Eq. \eqref{eq:vlb} encourages the posterior $q(\boldsymbol{z}|\boldsymbol{x})$ to resemble the age-specific prior $p(\boldsymbol{z}|c)$. This is the main mechanism for linking latent representations with age prediction: on the one hand, latent representations generated from the predicted $c$ have to resemble the latent representation of the input image and on the other hand, age-linked variation in the latent space is encouraged to follow a direction defined by $\boldsymbol{u}$. We used the SGVB estimator with the reparametrization trick \cite{Kingma13} to optimize the expectation in the last two terms of Eq. \eqref{eq:vlb}. 

Lastly, it has been shown that the supervised training of end-to-end feed-forward neural networks often suffers from over-fitting problems, whereas unsupervised autoencoders can often learn robust and meaningful intermediate features that are transferable to supervised tasks \cite{Zhuang15}. By combining the two frameworks, our model therefore allows for the sharing of low-level features (e.g., by convolutional layers) jointly learned by the autoencoder and regressor. 

\section{Experiments}
Understanding structural changes of the human brain as part of normal aging is an important topic in neuroscience. One emerging approach for such analysis is to learn a model that predicts age from brain MR images and then to interpret the patterns learned by the model. 
We tested the accuracy of the proposed regression model in predicting age from MRI based on two  implementations\footnote{Implementation based on \texttt{Tensorflow 1.7.0, keras 2.2.2}. Source code available at https://github.com/QingyuZhao/VAE-for-Regression}: the first implementation was based on a multi-layer perceptron neural network (all densely connected layers) applied to ROI-wise brain measurements while the second implementation was based on convolutional neural networks (CNN) applied to 3D volume images focusing on the ventricular area. Both implementations were cross-validated on a dataset consisting of T1-weighted MR images of 245 healthy subjects (122/123 women/men; ages 18 to 86) \cite{Adeli2018}. There was no group-level age difference between females and males (\textit{p} = 0.51, two-sample \textit{t}-test). 

\subsection{Age Prediction Based on ROI Measures}
With respect to the perceptron neural network, the input of the encoder were the z-scores of 299 ROI measurements generated by applying FreeSurfer (V 5.3.0) to the skull-stripped MR image of each subject \cite{Adeli2018}. The measurements consisted of the mean curvature, surface area, gray matter volume, and average thickness of 34 bilateral cortical ROI, the volumes of 8 bilateral subcortical ROIs, the volumes of 5 subregions of the corpus callosum, the volume of all white matter hypointensities, the left and right lateral and third ventricles, and the supratentorial volume (svol). 

The input to the encoder was first densely connected to 2 intermediate layers of dimension (128,32) with \texttt{tanh} as the activation function. The resulting feature was then separately connected to two layers of dimension 8 yielding the mean and diagonal covariance of the latent representation. The regressor shared the 2 intermediate layers of the encoder (Fig. \ref{fig:formulation}b) and produced the mean and standard deviation for the predicted age. The decoder took a latent representation as input and adopted an inverse structure of the encoder for reconstruction. 

\subsection{Age Prediction Based on 3D MR Images}
Taking advantage of recent advances in deep learning, the second implementation was build upon convolutional neural networks that directly took 3D images as input. All skull-stripped T1 images were registered to the SRI24 atlas space and down-sampled to 2mm isotropic voxel size. Since it has been well established that the ventricular volume significantly increases with age \cite{Kaye92}, we then tested if the model would learn this structural change in predicting age. To do this, each image was cropped to a 64*48*32 volume containing the ventricle region and was normalized to have zero mean and unit variance. This smaller field of view allowed for faster and more robust training of the following CNN model on limited sample size (N=245). Specifically, the encoder consisted of 3 batches of 3*3*3 convolutional layers with rectified linear unit (ReLU) activation and 2*2*2 max pooling layers. The sizes of feature banks in the convolutional layers were (16,32,64) respectively. 

Similar to the previous implementation, the extracted features were fed into 2 densely connected layers of dimension (64,32) with \texttt{tanh} activation function. The final dimension of latent space was 16. The regressor shared the convolutional layers of the encoder and also had 2 densely connected layers of (64,32). The decoder had an inverse structure of the encoder and used \texttt{Upsampling3D} as the inverse operation of max pooling. Since the CNN-based implementation had substantially more model parameters to determine than the first implementation, L2 regularization was applied to all densely connected layers. 

\subsection{Measuring Accuracy}

The accuracy of each implementation was reported based on a 5-fold cross-validation measuring the R2 score (coefficient of determination, the proportion of the variance in age that is predictable from the model) and root mean squared error (rMSE). The outcome of each approach was compared to 7 other regression methods, of which 6 were non-neural-network methods as implemented in \textit{scikit-learn} 0.19.1: linear regression (LR), Lasso, Ridge regression (RR), support vector regression (SVR), gradient-boosted tree (GBT), k-nearest neighbour regression (K-NN). The last approach was a single neural-network regressor (NN), i.e., the component corresponding to $q(c|\boldsymbol{x})$ without outputting standard deviation. 

With respect to the ROI-based experiments, optimal hyperparameters of the scikit-learn methods (except for LR) were determined through a 10-fold inner cross-validation (an overall nested cross-validation). Specifically, we searched $C\in\{1,10,...,10^3\},\gamma\in\{10^{-2},...,10^2\}$ for SVR,  $N\in\{10,50,100,500\}$ for GBT, $\alpha\in\{10^{-3},...,10^4\}, \gamma\in\{10^{-2},...,10^2\}$ for RR, and $N\in\{1,5,10,50\}$ for K-NN. 

With respect to the 3D-image-based experiments, nested cross-validation was extremely slow for certain methods (e.g. GBT, CNN), so we simply repeated the outer 5-fold cross-validation using the hyperparameters defined in the above search space and reported the best accuracy. The search space of the L2 regularization for NN and our method was \{0, .001, .01, .1, 1\}. 

\begin{table}[!t]
\centering

\caption{Age prediction accuracy of different methods based on ROI measurements and 3D volume images of ventricle.}
\begin{tabular}{ c | c || p{1.2cm} | p{1.2cm} | p{1.2cm} |p{1.2cm} |p{1.2cm} | p{1.2cm}| p{1.3cm} } 
 \hline
 & & LR & RR & SVR & GBT & K-NN & NN &   \textbf{Ours} \\ 
 \hline\hline
 \multirow{2}{4.8em}{ROI Measures} & R2 & 0.107 & 0.336 & 0.311 & 0.64 & 0.535 & 0.563  & \textbf{0.666} \\ 
 & rMSE & 14.6  &  12.6 & 12.8  &  9.3 & 10.5  & 10.3 & \textbf{9.0} \\ 
 \hline\hline
 \multirow{2}{4.8em}{3D Volume} &  R2 & 0.737 & 0.737 &  0.737 & 0.719 & 0.549  & 0.79  & \textbf{0.808}\\ 
 & rMSE & 7.8  & 7.8  & 7.8  & 8.2  &  10.5   & 7.0  & \textbf{6.9} \\ 
 \hline
\end{tabular}
\label{tb:acc}
\end{table}

\subsection{Results}
As Table \ref{tb:acc} shows, age prediction based on 3D images of ventricle was generally more accurate than on ROI measurements. The two neural-network-based predictions were the most accurate in terms of R2 and rMSE. Fig. \ref{fig:pred} shows the predicted age (in the 5 testing folds) estimated by our model versus ground-truth. The best prediction was achieved by our model applied to the 3D ventricle images, which yielded a 6.9-year rMSE. In the ROI-based experiment, our model was more accurate than the single neural-network regressor (NN), which indicates the integration of VAE for modeling latent representations could regularize the feed-forward regressor network. In the 3D-image-based experiment, our model was more accurate than NN either with (Table \ref{tb:acc}) or without L2 regularization (our model and NN yielded R2 of 0.761 and 0.745 respectively). Even though this improvement was not as significant as in the ROI-based experiment, our model enabled direct visualization of brain developmental patterns.

\begin{figure}[!tb]
	\centering
    \includegraphics[width=0.9\linewidth]{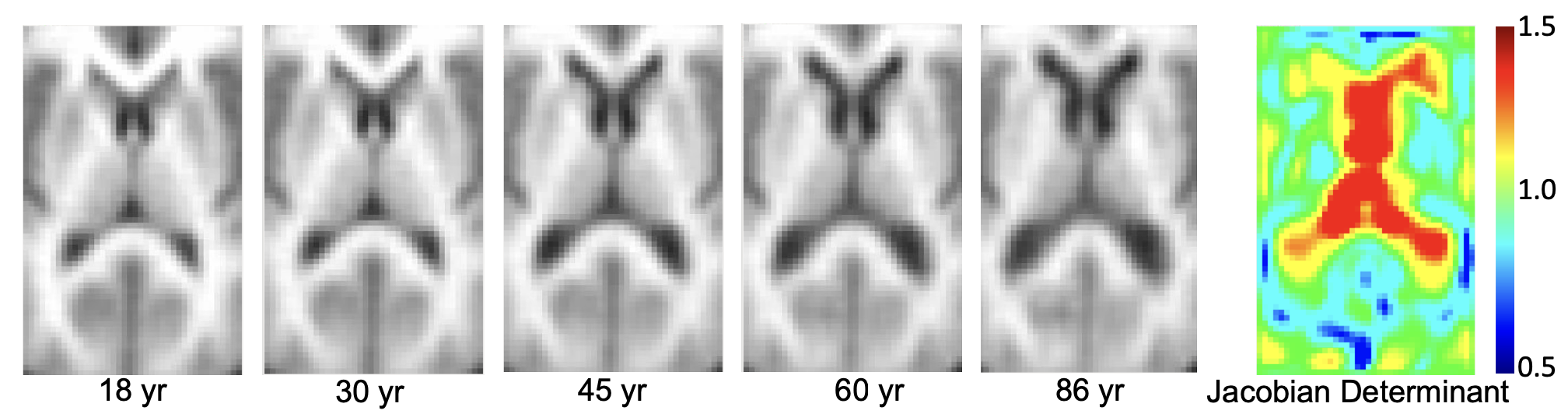}
	\caption{Left: Brain images reconstructed from age-specific latent representations. Right: Jacobian determinant map derived from the registration between the 18 year old brain and the 86 year old brain. The major expanding region is located on the ventricle.}
	\label{fig:ventricle}
\end{figure}

\begin{figure}[!tb]
	\centering
    \includegraphics[width=1\linewidth]{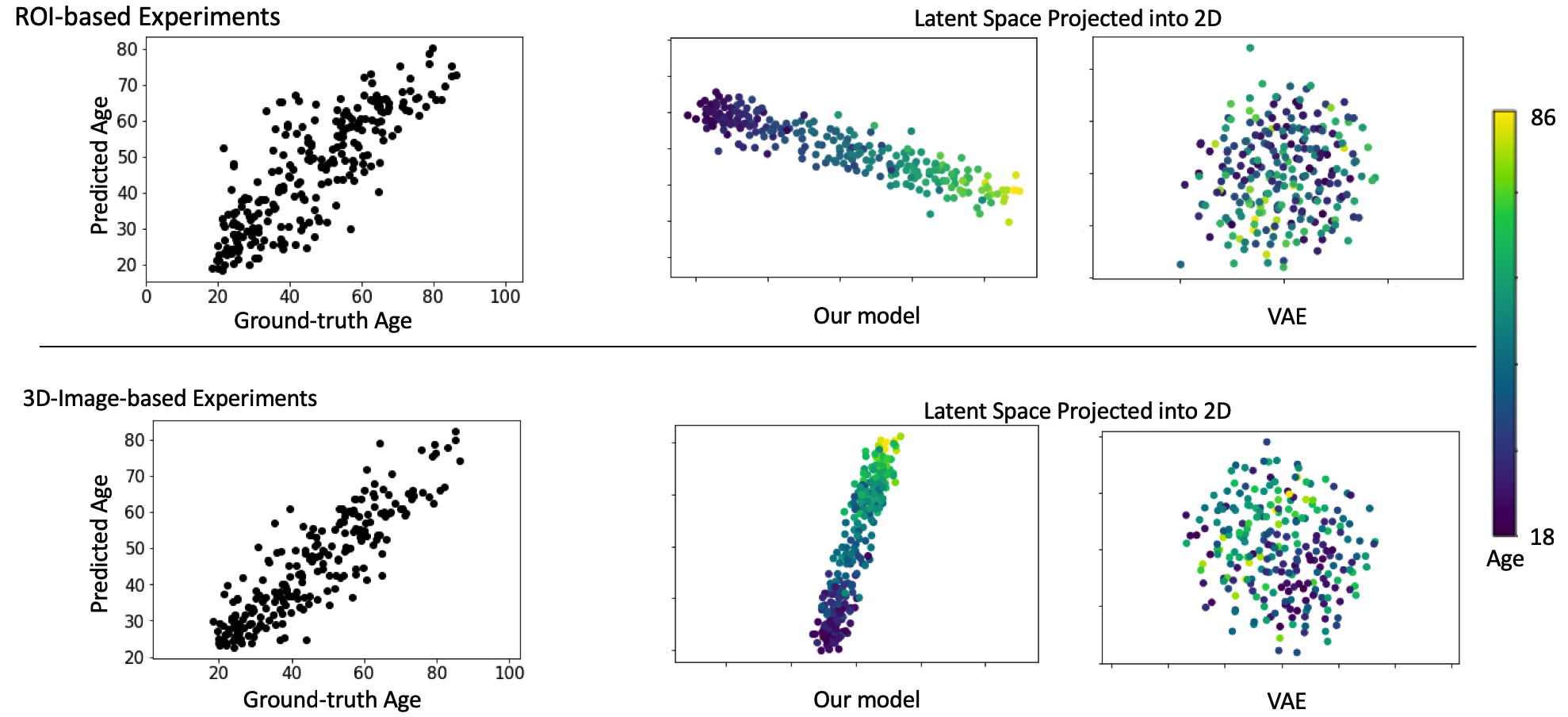}
	\caption{Upper row: results of ROI-based experiments. Lower row: results of 3D-image-based experiments. Left: Predictions made by our model vs. ground-truth. Middle: Latent representations estimated by our model. Right: Latent representations estimated by traditional VAE.}
	\label{fig:pred}
\end{figure}

Indeed, despite the tremendous success of deep learning in various applications, interpretability of the black-box CNN (e.g., which input variable leads to accurate prediction, or what specific features are learned) remains an open research topic. Most existing solutions can only produce a ``heat map" indicating the location of voxels that contribute to faithful prediction, but this does not yield any semantic meaning of the learned features that can improve mechanistic understanding of the brain. Thanks to the generative modelling, our formulation provides an alternative way for interpreting the aging pattern captured by the CNN. Specifically, Fig. \ref{fig:ventricle} shows the simulated ``mean brain images" at different ages by decoding age-specific latent representations $\{\boldsymbol{z}=\boldsymbol{u}^{\textbf{T}}c|c\in[18,86]\}$, i.e., mean of the latent generator $p(\boldsymbol{z}|c)$. We can clearly observe that the pattern learned by the model for age prediction was mainly linked to the enlargement of ventricle. This result is consistent with current understanding of the structural development of the brain.

Lastly, we show in Figure 3 that the dimension related to age was disentangled from the latent space. In both ROI-based and image-based experiments, we trained our model on the entire dataset. The resulting latent representations were transformed from the latent space to a 2D plane via TSNE and color-coded by the ground-truth age. We observe that one direction of variation is associated with age, whereas the unsupervised training of traditional VAE does not lead to clear disentanglement. 

\section{Conclusion and Discussion}
In this paper, we introduced a generic regression model based on the variational autoencoder framework and applied it to the problem of age prediction from structural MR images. The novel generative process enabled the disentanglement of age as a factor of variation in the latent space. This did not only produce more accurate prediction than a regular feed-forward regressor network, but also allowed for synthesizing age-dependent brains that facilitated the identification of brain aging pattern. Future direction of this work includes simultaneously disentangling more demographics factors of interest, e.g. sex, disease group, to study compounding effects, e.g. age-by-sex effects or accelerated aging caused by disease.  

\textbf{Acknowledgements} This research was supported in part by NIH grants U24AA021697, AA005965, AA013521, AA017168, AA026762

\bibliographystyle{splncs}
\bibliography{references}
\end{document}